\documentclass[runningheads]{llncs}
\usepackage{graphicx}
\usepackage{subcaption}
\usepackage{tikz}
\usepackage{caption}
\usepackage{subcaption}
\usepackage{tikz}
\usepackage{pgfplots}
\usetikzlibrary{spy,calc}
\usepackage{hyperref}
\usepackage{verbatim}
\usepackage{booktabs,array} 
\usepackage{url}

\makeatletter
\newcommand{\thickhline}{%
    \noalign {\ifnum 0=`}\fi \hrule height 1pt
    \futurelet \reserved@a \@xhline
}

\begin{document}
\title{A Divide-and-Conquer Approach Towards Understanding Deep Networks}
%
%
\author{Weilin Fu\inst{1}
\and Katharina Breininger\inst{1}
\and Roman Schaffert\inst{1}
\and Nishant Ravikumar \inst{1}
\and Andreas Maier\inst{1, 2}}

%
\authorrunning{Fu et al.}
%
\institute{Pattern Recognition Lab, Friedrich-Alexander University Erlangen-N\"urnberg, 91058 Erlangen, Germany \and
Erlangen Graduate School in Advanced Optical Technologies(SAOT), 91058 Erlangen, Germany\\
\email{weilin.fu@fau.de}}


%
\maketitle              
\begin{abstract}
\textcolor{black}{Deep neural networks have achieved tremendous success in various fields including medical image segmentation. However, they have long been criticized for being a black-box, in that interpretation, understanding and correcting architectures is difficult as there is no general theory for deep neural network design. Previously, precision learning was proposed to fuse deep architectures and traditional approaches. Deep networks constructed in this way benefit from the original known operator, have fewer parameters, and improved interpretability. However, they do not yield state-of-the-art performance in all applications. }
\textcolor{black}{In this paper, we propose to analyze deep networks using known operators, by adopting a divide-and-conquer strategy to replace network components, whilst retaining its performance. The task of retinal vessel segmentation is investigated for this purpose. We start with a high-performance U-Net and show by step-by-step conversion that we are able to divide the network into modules of known operators. The results indicate that a combination of a trainable guided filter and a trainable version of the Frangi filter yields a performance at the level of U-Net (AUC 0.974 vs. 0.972) with a tremendous reduction in parameters (111,536 vs. 9,575). In addition, the trained layers can be mapped back into their original algorithmic interpretation and analyzed using standard tools of signal processing.}

\end{abstract}

\section{Introduction}
\textcolor{black}{Deep learning (DL) technology~\cite{lecun2015deep} has been successfully applied in various fields including medical image segmentation, which provides substantial support for diagnosis, therapy planning and treatment procedures. Despite its outstanding achievements, DL-based algorithms have long been criticized for being a black-box and many design choices in Convolutional Neural Network (CNN) topologies are driven rather by experimental improvements than theoretical foundation. Accordingly, understanding the actual working principle of the architectures is difficult. One option to gain interpretability is to constrain the network with known operators. Precision learning~\cite{maier2018precision,maier2019learning}, which integrates known operators~\cite{fu2018frangi,wurfl2016deep} into DL models, can provide a suitable mechanism to design CNN architectures. This strategy integrates prior knowledge into the deep learning pipeline, thereby improving interpretability, providing guarantees and quality control in certain settings. However, the quantitative performance of these approaches often falls short compared to completely data-driven approaches.} 

\textcolor{black}{In this work, we propose an approach to debug and identify the limitation/bottleneck of a known operator workflow. Frangi-Net~\cite{fu2018frangi}, which is the deep learning counterpart of the Frangi filter~\cite{frangi1998multiscale} is utilized as an exemplary network. Performance of different methods are evaluated on the retinal vessel segmentation task, using data from the Digital Retinal Images for Vessel Extraction (DRIVE) database~\cite{staal2004ridge}. Experiments are designed under the assumption that if the replacement of one step leads to a performance boost, then this step is the probable bottleneck of the overall workflow. In our case, we debug the Frangi-Net by replacing the preprocessing step with the powerful U-Net~\cite{ronneberger2015u}. With the output from the U-Net as input, Frangi-Net approaches the state-of-the-art performance. Thereby, we conclude that the preprocessing method is the weakness of the Frangi-Net segmentation pipeline. In other words, given a proper preprocessing algorithm, Frangi-Net may be capable of accomplishing the retinal vessel segmentation task. To verify this hypothesis, we further utilize the guided filter layer~\cite{wu2018fast}, which is a deep learning module designed for image quality enhancement. Experimental results confirm our hypothesis: the additional guided filter layer indeed brings about a substantial improvement in performance. \textcolor{black}{Due to the modular design, analysis of the trained filter block is possible which reveals slightly unexpected behaviour.} Our work has two main contributions: Firstly, we propose a feasible way to identify the bottleneck of a precision learning-based workflow. 
Secondly, the debugging procedure yields a network pipeline with well-defined explainable steps for retinal vessel segmentation, i.\,e.\, guided filter layer for preprocessing, and Frangi-Net for vesselness computation.
}

\section{Methods}
\subsection{Frangi-Net}
\begin{figure}[t]
  \centering
    \includegraphics[width=0.95\textwidth]{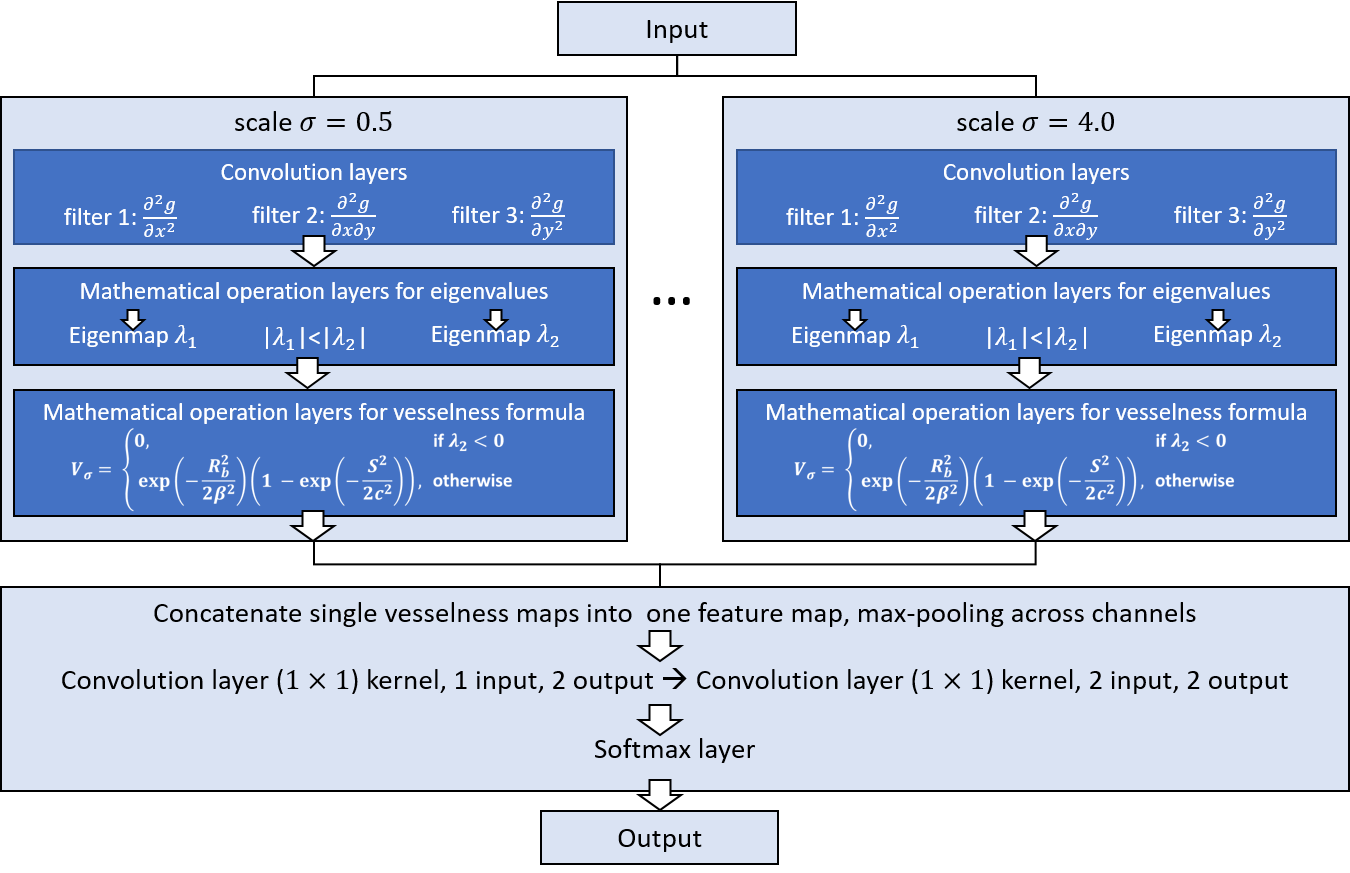}
    \caption{Architecture of the 8-scale Frangi-Net.}
    \label{fig:frangi}
\end{figure}

\textcolor{black}{In this work, Frangi-Net, which is the deep learning counterpart of the Frangi filter, is utilized as the segmentation network in different pipelines. The Frangi filter is a widely used multi-scale tube segmentation method, which calculates vesselness response $V_0$ of dark tubes at scale $\sigma$ with Hessian eigenvalues ($|\lambda_1| \leq |\lambda_2|$) using:\begin{equation} \label{eqn:Frangi}
V_0(\sigma) = \left\{\begin{array}{ll}
0, & \textrm{ if } \lambda_2 < 0,\\ 
\exp(-\frac{R_B^{2}}{2\beta ^{2}})(1-\exp(-\frac{S^2}{2c^2})), & \textrm{ otherwise,}
\end{array}
\right.
\end{equation} where $S = \sqrt{\lambda_1^2 + \lambda_2^2}$ is the second-order structureness, $R_B = \frac{\|\lambda_1\|}{\|\lambda_2\|}$ is the blobness measure, and $\beta, c$ are image-dependent parameters for blobness and structureness terms. The Frangi-Net is constructed by representing each step in the multi-scale Frangi filter as a layer. Here, we employ a Frangi-Net with 8 different Gaussian scales ranging from 0.5 to 4.0. The convolution kernels are initialized as the second-order partial derivatives of the Gaussian kernel at the corresponding scales. We employ two additional $1\times 1$ convolution layers before the final softmax output layer, to regulate the data range. The hyper-parameters $\beta, c$ in Eq.~\ref{eqn:Frangi} of all scales are initialized to 0.5 and 1.0, respectively. The network has 6,\,525 weights, and the overall architecture is shown in Fig.~\ref{fig:frangi}. }

\subsection{U-Net}
\textcolor{black}{In this work, a U-Net is directly applied to retinal vessel segmentation, and forms the baseline method for all comparisons. U-Net is a successful encoder-decoder CNN architecture, popularized in the field of medical image segmentation. It combines location information in the contracting encoder path, with contextual information in the expanding decoder path via skip connections. Here, we adapt a three-level U-Net with 16 initial features with two main modifications. Firstly, batch normalization layers are added after convolution layers to stabilize the training process. Secondly, deconvolution layers are replaced with upsampling layers followed by a $1\times 1$ convolution layer. The overall architecture contains 111,536 trainable weights.}

\subsection{U-Net + Frangi-Net}

\begin{figure}[t]
  \centering
    \includegraphics[width=0.65\textwidth]{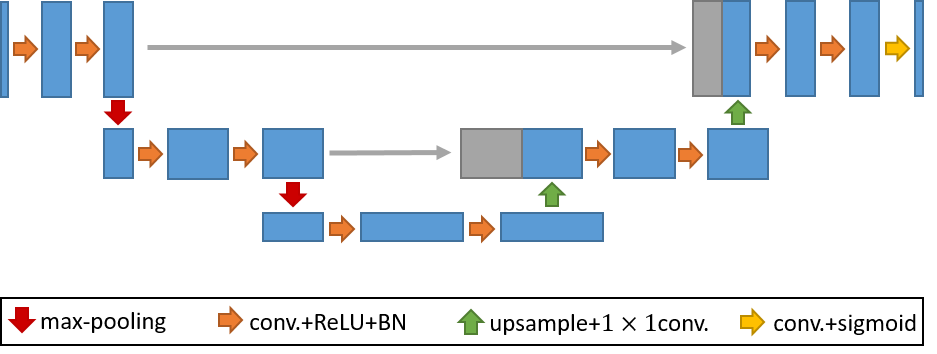}
    \caption{U-Net architecture adapted for preprocessing.}
    \label{fig:unet}
\end{figure}

\textcolor{black}{In order to analyze the reason for the performance differences between Frangi-Net and the U-Net, we propose to employ the latter as a ``wildcard preprocessing network". To this end, we concatenate the two networks such that the output of the U-Net serves at input for the Frangi-Net and train the segmentation pipeline end-to-end. The intuition here is that, if the combined network is able to achieve a performance on par with the completely data driven approach, the bottleneck of the known-operator network lies in the preprocessing. Otherwise, the known operator is inadequate to solve the task at hand, even with optimized images. Since Frangi-Net only takes single channel input, two additional modifications are made to the final layers of U-Net: the final convolution layer yields a one channel output, and a sigmoid layer is employed to replace the softmax layer for feature map activation. The modified U-Net architecture is shown in Fig.~\ref{fig:unet}.}

\subsection{Guided Filter Layer + Frangi-Net}

\begin{figure}[]
  \centering
    \includegraphics[width=0.6\textwidth]{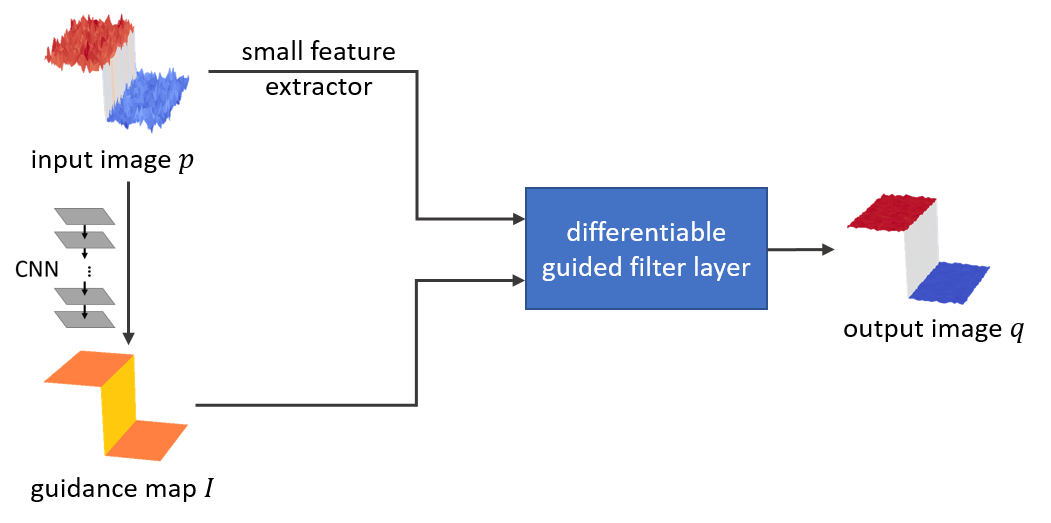}
    \caption{Architecture of guided filter layer adapted for preprocessing.}
    \label{fig:gf_layer}
\end{figure}

\textcolor{black}{Preliminary results for experiments conducted using U-Net and U-Net + Frangi-Net indicated that the preprocessing step was indeed the bottleneck in the vessel segmentation pipeline. Consequently, we propose to replace the ``wildcard'' U-Net with a guided filter layer. }\textcolor{black}{The guided filter layer was proposed as differentiable neural network counterpart of the guided filter~\cite{he2013guided}, which can be utilized as an edge-preserving, denoising approach. The guided filter takes one image $p$ and one guidance image $I$ as input to produce one output image $q$. This translation-variant filtering process can be simplified and described in Eq.~\ref{eq:gf}:\begin{equation}\label{eq:gf}q_i = \sum_j W_{ij} (I)p_j,\end{equation} where $i, j$ are pixel indices, and $W_{ij}$ is the kernel which is a function of the guidance image $I$ and is independent of $p$.}
\textcolor{black}{
A guided filter layer with two trainable components is used as the preprocessing block. First, the guidance map $I$ is generated with a CNN, using image $p$ as input. Here the CNN is configured as a five-layer Context Aggregation Network (CAN)~\cite{chen2017fast}. Subsequently, a small feature extractor is applied to image $p$ before being passed to the guided filter layer. This feature extractor is composed of two $3\times 3$ convolution layers with five intermediate, and one final output feature maps. The guided filter block contains 3,\,050 parameters, and the architecture is shown in Fig.~\ref{fig:gf_layer}.}

\section{Experiments and Results}
\subsection{Data}
\textcolor{black}{The DRIVE database is employed to evaluate different pipelines in this study. The database contains 40 RGB fundus photographs of size $565\times 584$ pixels, which are evenly divided into a training and a testing set. A validation set of four images is further separated from the training set to monitor the training process and avoid over-fitting.}
\textcolor{black}{The green channels, which have the best contrast between vessels and the background, are extracted and processed using Contrast Limited Adaptive Histogram Equalization (CLAHE) to balance the inhomogeneous illumination.}
Manual labels and Field Of View (FOV) masks are provided for all images. For each image of the training set, a weighting map $w$ which emphasizes thin vessels is generated on the basis of the manual label using the equation $w=\frac{1}{\alpha \times d}$, where $d$ denotes the vessel diameter in the ground truth, and $\alpha$ is a factor manually set to $0.18$. In order to have a meaningful and fair comparison between different methods, all FOV masks are eroded inward by four pixels to remove potential border effects. Performance evaluation is conducted inside the FOV masks. 

\subsection{Network Training}
\textcolor{black}{The objective functions for all learning-based methods in this work are constructed with three parts as: $L_{total} = w\cdot L_{focal} + \lambda_w\cdot R_{w} + \lambda_s\cdot R_s$, where, $w$ is the weighting map which emphasizes small vessels; $L_{focal}$ is the class balanced focal loss~\cite{lin2017focal}, with a focusing factor of 2.0; $R_w$ denotes an $\ell_2$-norm regularizer on the network weights to prevent overfitting; $R_s$ represents a similarity regularizer which is the mean squared error between the input and output of the preprocessing net. $\lambda_w, \lambda_s$ are the scaling factors of the corresponding regularizers, and are set to $0.2$ and $0.1$, respectively. The Adam optimizer~\cite{kingma2014adam} with learning rate decay is utilized to minimize the objective function. The initial learning rate is $5\times 10^{-4}$ for U-Net, and $5\times 10^{-5}$ for all other pipelines. All networks are trained with a batch size of 50, and with $168\times168$ image patches. Data augmentation in form of rotation, shearing, Gaussian noise, and intensity shifting is employed. }\textcolor{black}{All methods are implemented in Python 3.5.2 using TensorFlow 1.10.0.}

\subsection{Evaluation and Results}

\begin{figure}[t]
    \centering
    \begin{subfigure}[b]{0.24\textwidth}
        \centering
        \includegraphics[width=\textwidth]{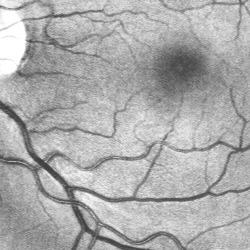}
    \end{subfigure}
    \hfill
    \begin{subfigure}[b]{0.24\textwidth}
    \centering
    \includegraphics[width=\textwidth]{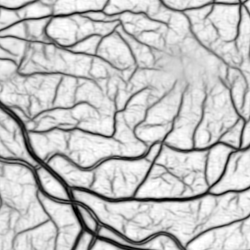}
    \end{subfigure}
    \hfill
    \begin{subfigure}[b]{0.24\textwidth}
    \centering
    \includegraphics[width=\textwidth]{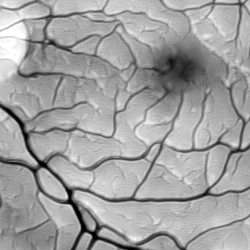}
    \end{subfigure}
    \hfill
    \begin{subfigure}[b]{0.24\textwidth}
    \centering
    \includegraphics[width=\textwidth]{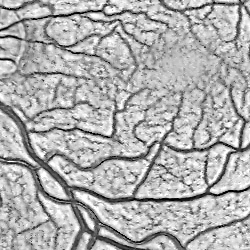}
    \end{subfigure}
    \\
    \begin{subfigure}[b]{0.24\textwidth}
        \centering
        \includegraphics[width=\textwidth]{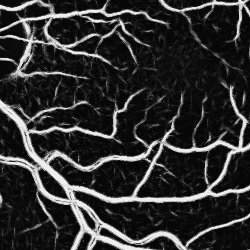}
        \caption{}
    \end{subfigure}
    \hfill
    \begin{subfigure}[b]{0.24\textwidth}
    \centering
    \includegraphics[width=\textwidth]{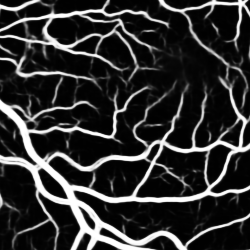}
    \caption{}
    \end{subfigure}
    \hfill
    \begin{subfigure}[b]{0.24\textwidth}
    \centering
    \includegraphics[width=\textwidth]{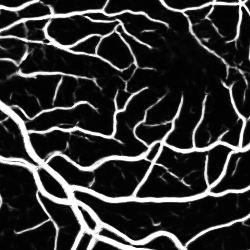}
        \caption{}
    \end{subfigure}
    \hfill
    \begin{subfigure}[b]{0.24\textwidth}
    \centering
    \includegraphics[width=\textwidth]{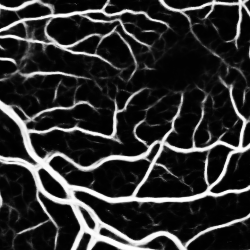}
        \caption{}
    \end{subfigure}
    \caption{The input (a) and output (b-d) of preprocessing networks for a representative ROI are shown in the upper row, the corresponding probability map results after Frangi-Net are presented in the lower row for: (a) no preprocessing network, (b) UP, (c) UP with $R_s$, (d) GF. }\label{fig:result}
\end{figure}

The evaluation performance of 6 different segmentation workflows is evaluated on the DRIVE testing set, and is summarized in Table~\ref{tab:result_small}. Binarization of the output probability maps from the network pipelines is performed with a single threshold which maximizes the F1 score on the validation set. The input, intermediate outputs of the preprocessing nets and the corresponding probability map results from the Frangi-Net for an representative region of interest (ROI) of an image from the testing set are presented in Fig.~\ref{fig:result}.


From Table~\ref{tab:result_small}, we observe that the Frangi-Net without additional preprocessing (FN) performs better than the original Frangi filter (FF), but worse than the completely data-driven U-Net (UN). Using the U-Net as a preprocessing network (UP + FN), we observe a performance boost, achieving results on-par with UN, with respect to all evaluation metrics and reaching an AUC score of $0.975$. With an additional regularizer $R_s$ that enforces the similarity between the input and output of the preprocessing network, the performance is only modestly impaired. When looking at the intermediate outputs of the preprocessing nets (see Fig.~\ref{fig:result} (b) and (c)), we observe that the UP substantially enhances the contrast for small vessels and reduces noise compared to the input image (a). Low frequency information, e.\,g., the illumination in the bright optic disc and the dark macula region, is removed when no additional $R_s$ is applied.
This provides further confirmation of the hypothesis that the main bottleneck of the proposed known-operator pipeline lies in the preprocessing, and can be combated by an appropriate adaption of this step. This is supported by the results achieved using the guided filter layer for preprocessing (GF + FN). 
\textcolor{black}{
The guided filter layer, however, does not simply learn an edge-preserving denoising filtering as the intermediate output reveals (see Fig.~\ref{fig:result} (d)). It performs a substantial enhancement of small vessels and removal of the low-frequency background comparable to UP (see Fig.~\ref{fig:result} (b)). In this case, the performance of the pipeline is only marginally inferior to that of the U-Net, approaching an AUC score of 0.972.}


\begin{table}[t]
\centering
\caption{Performance evaluation on DRIVE testing set. \textit{prep.}, \textit{reg.}, \textit{seg.} denote \textit{preprocessing net}, \textit{regularizer}, and \textit{segmentation method}, respectively.} 
\vspace{.2cm}
\label{tab:result_small}
\begin{tabular}{|r|r|r|ccccc|}
\thickhline
\multicolumn{1}{|c|}{prep.} & \multicolumn{1}{c|}{reg.} & \multicolumn{1}{c|}{seg.} & specificity   & sensitivity   & F1 score      & accuracy      & AUC    \\ \thickhline
-     & -    & FF   & .9616$\pm$.0150   & .7528$\pm$.0612   & .7477$\pm$.0323   & .9341$\pm$.0089   & .9401  \\ \hline
-     & -    & FN   & .9633$\pm$.0125   & .8008$\pm$.0590   & .7812$\pm$.0256   & .9419$\pm$.0070   & .9610  \\ \hline
-     & $R_w$    & UN   & .9756$\pm$.0057   & .7942$\pm$.0576   & .8097$\pm$.0227   & .9516$\pm$.0056   & .9743  \\ \thickhline
\thickhline
UP    & $R_w$    & FN   & .9726$\pm$.0082 & .8070$\pm$.0565 & .8088$\pm$.0236 & .9506$\pm$.0054 & .9743 \\ \hline
UP    & $R_w, R_s$  & FN   & .9753$\pm$.0057 & .7715$\pm$.0598 & .7949$\pm$.0248 & .9485$\pm$.0060 & .9703 \\ \hline
GF    & -    & FN   & .9729$\pm$.0060 & .7982$\pm$.0546 & .8048$\pm$.0191 & .9498$\pm$.0048 & .9719 \\ \thickhline
\end{tabular}
\end{table}

\section{Discussion and Conclusion}
\textcolor{black}{
We proposed a method to analyze and interpret a DL-based algorithm, via step-by-step conversion of a fully-data driven approach, to construct a pipeline using well-defined known operators. The approach helps to identify and combat bottlenecks in a known-operator pipeline, by localizing the components responsible for drops in performance. Additionally, it provides a mechanism to interpret deep network architectures in a divide-and-conquer pattern, by replacing each step in the network pipeline with a well-defined operator.}

\textcolor{black}{
The potential of the proposed framework to improve our understanding of deep neural networks and enable intelligent network design was demonstrated for the exemplary task of retinal vessel segmentation. The previously proposed known-operator network Frangi-Net enables easy interpretation, but performs worse than a fully data-driven approach such as the U-Net. Conversely, an interpretation of the fully data-driven approach remains vague despite satisfactory performance. 
By using the U-Net as a debugging tool, we confirm that with appropriate preprocessing, the Frangi-Net is capable of achieving on-par performance. This performance boost also indicates that the preprocessing is the bottleneck of the Frangi-Net workflow. 
Subsequently, we identify the guided filter layer as a suitable known operator that can serve as a replacement for the U-Net in terms of preprocessing, while retaining performance. }

\textcolor{black}{
The quantitative results support our hypothesis that the task of vessel segmentation can be separated into two steps: a preprocessing step that enhances image quality, and a segmentation step which yields the actual vesselness probability map. By replacing these elements step-by-step, we are able to preserve high segmentation performance while incorporating interpretability into the network pipeline with well-defined, understandable steps. }

\textcolor{black}{While the results of the U-Net preprocessing with similarity regularization demonstrate that there exists an edge-preserving filtering approach that results in an equally effective segmentation based on the vesselness filter, the guided filter layer does not fulfill the expected filtering behavior. Instead of edge-preserving filtering, the guided filter layer learns a domain transfer to an vessel-enhancing representation that removes low frequency information at the same time. Looking at Eq.~\ref{eq:gf} this seems surprising, as the guided filter uses the guidance image only for design of the filter kernel in a shift-variant filtering process. Yet, this design does not guarantee an edge-preserving filtering per se as the guidance image may also result in band-pass kernels. As a result, the filter learns to create kernels that are optimal with respect to the purpose of the net that is a vessel enhanced image in our case.}

\textcolor{black}{Still, our divide-and-conquer approach allows to specify the important parts of a network. This is achieved by showing that a known operator network which is restricted in what it can learn with 9,\,575 vs. 111,\,536 parameters, performs comparably to a completely data-driven network with an AUC score of 0.972 vs. 0.974. The use of a powerful network, i.\,e., U-Net in this case, supplements the performance and addresses the shortcomings of the known operators, and thus helps to improve understanding of the network for a specific task. Future work will look into exploiting the divide-and-conquer approach to aid network interpretation and performance improvement for other tasks, based on known operator modules. It provides a systematic framework to design interpretable network pipelines with minimal loss in performance, relative to completely data-driven approaches, which is compelling for the intelligent design of networks in the future.}

%
%
%
\bibliographystyle{splncs04}
\bibliography{mybibliography}

\end{document}